\documentclass[letterpaper]{article} 
\usepackage{aaai2026} 
\usepackage{times} 
\usepackage{helvet} 
\usepackage{courier} 
\usepackage[hyphens]{url} 
\usepackage{graphicx} 
\usepackage{natbib} 
\usepackage{caption} 
\usepackage{amsmath}
\usepackage{amssymb}
\usepackage{amsfonts}
\usepackage{algorithm}
\usepackage{algorithmic}

\usepackage{newfloat}
\usepackage{listings}
\DeclareCaptionStyle{ruled}{labelfont=normalfont,labelsep=colon,strut=off} 
\floatstyle{ruled}
\newfloat{listing}{tb}{lst}{}
\floatname{listing}{Listing}
\title{MAPI-GNN: Multi-Activation Plane Interaction Graph Neural Network for\\Multimodal Medical Diagnosis}
\author {
    Ziwei Qin\equalcontrib,
    Xuhui Song\equalcontrib,
    Deqing Huang,
    Na Qin,
    Jun Li\protect\thanks{Corresponding author.}
}
\affiliations {
    Institute of Systems Science and Technology, School of Electrical Engineering, Southwest Jiaotong University, China    
    
    {qinziwei, sxh2023}@my.swjtu.edu.cn,
    {qinna, elehd}@swjtu.edu.cn,
    dirk.li@outlook.com
}

\begin{document} 

\maketitle

\begin{abstract}
Graph neural networks are increasingly applied to multimodal medical diagnosis for their inherent relational modeling capabilities. However, their efficacy is often compromised by the prevailing reliance on a single, static graph built from indiscriminate features, hindering the ability to model patient-specific pathological relationships. To this end, the proposed Multi-Activation Plane Interaction Graph Neural Network (MAPI-GNN) reconstructs this single-graph paradigm by learning a multifaceted graph profile from semantically disentangled feature subspaces. The framework first uncovers latent graph-aware patterns via a multi-dimensional discriminator; these patterns then guide the dynamic construction of a stack of activation graphs; and this multifaceted profile is finally aggregated and contextualized by a relational fusion engine for a robust diagnosis. Extensive experiments on two diverse tasks, comprising over 1300 patient samples, demonstrate that MAPI-GNN significantly outperforms state-of-the-art methods.
\end{abstract}

%
%
\begin{links}
    \link{Code}{https://github.com/HecateBlair/MAPI-GNN}
\end{links}


\section{Introduction}
\label{sec:intro}
Multimodal medical imaging is pivotal for comprehensive disease diagnosis, as it leverages diverse physical principles \cite{RN158, RN159, RN160}—such as capturing anatomical structure with MRI and metabolic activity with PET—to provide synergistic insights that are unattainable from any single modality \cite{RN161, RN162, RN163}. However, effectively fusing this intrinsically heterogeneous data remains a formidable challenge  \cite{RN167, RN171, RN172} . The core difficulty lies in reconciling disparate data structures and semantic information to form a cohesive pathological representation \cite{RN164, RN165}. While deep learning models, particularly Convolutional Neural Networks (CNNs)  \cite{RN173, RN174, RN175, RN176, RN177, RN207}, have advanced feature extraction, their inherent grid-based processing struggles to explicitly model the complex, high-dimensional, and often subtle non-Euclidean relationships between different modalities \cite{RN168, RN169, RN170}.
This limitation has clinical ramifications: failing to capture complex cross-modal patterns can obscure pathological markers, compromising accuracy and clinical outcomes.

\begin{figure}[t]
\centering
\includegraphics[width=1\columnwidth]{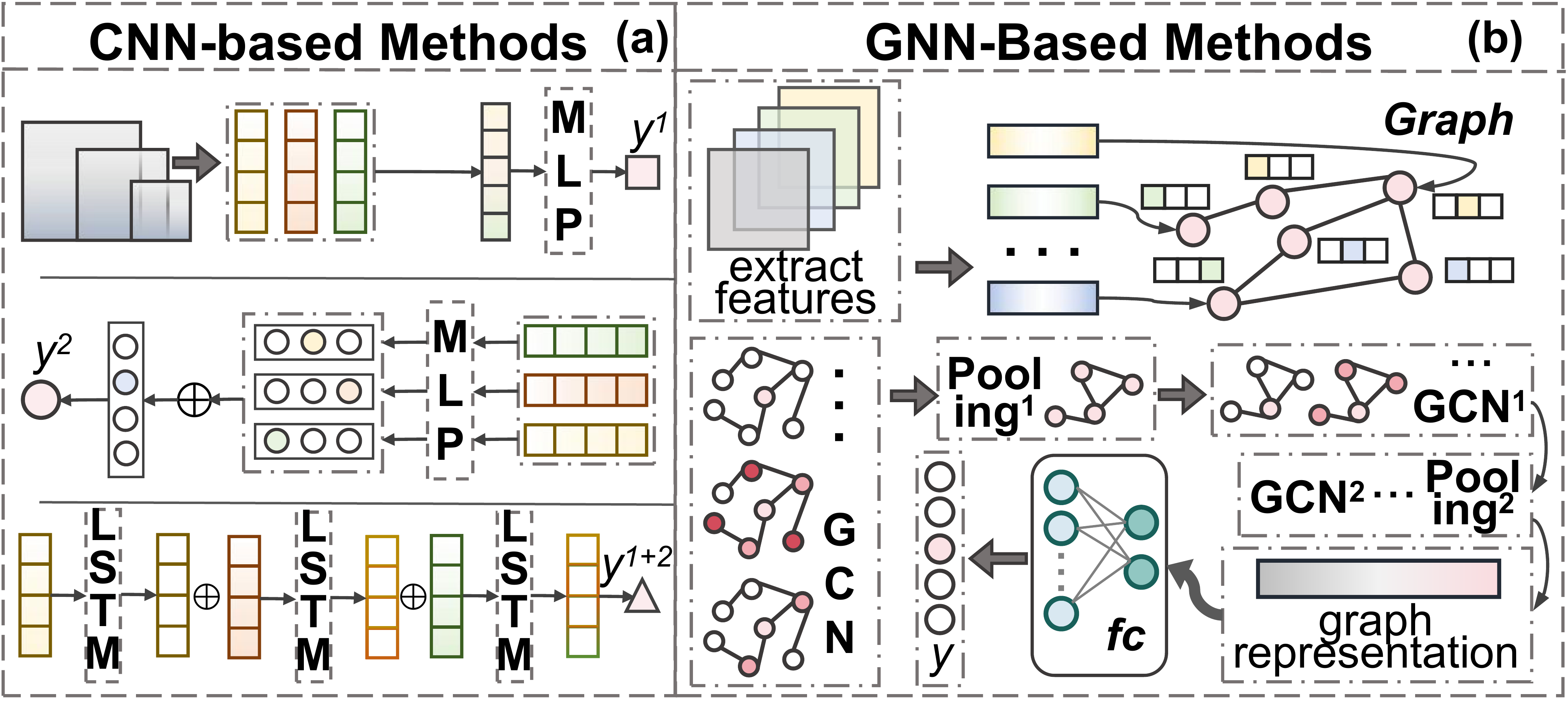}
\caption{Conceptual fusion strategies. (a) CNN-based methods use fixed fusion points (e.g., early/late), which limits the modeling of complex inter-modal dependencies. (b) GNNs offer a flexible paradigm, using a graph topology for explicit relationship modeling and hierarchical aggregation.}
\label{fig:fig2}
\end{figure}

\begin{figure*}[t]
\centering
\includegraphics[width=0.88\linewidth]{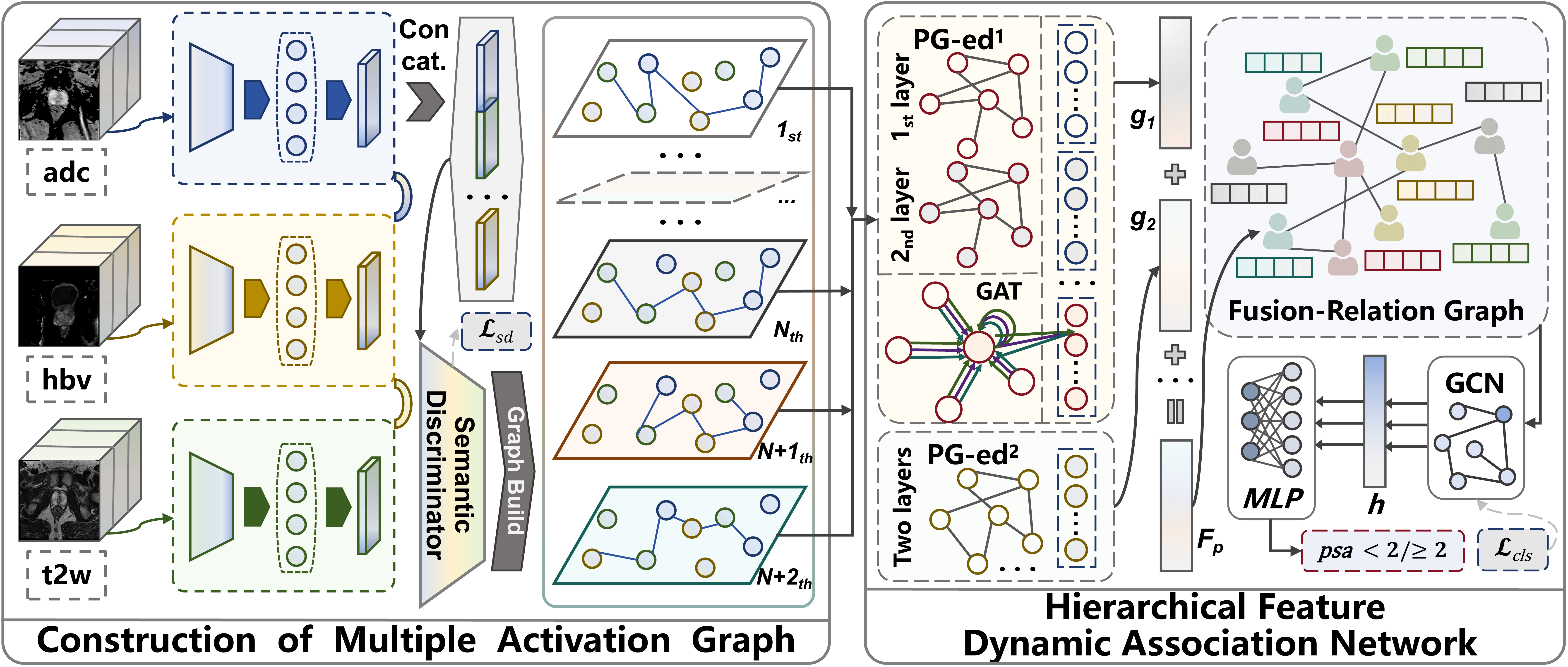}
\caption{Overview of the MAPI-GNN architecture. Stage I (detailed in Fig. \ref{fig:fig4}) generates multiple, semantically-aware activation graphs from patient-specific multimodal data. Stage II (detailed in Fig. \ref{fig:fig5}) then performs a hierarchical, two-level fusion on these graphs, first modeling intra-sample relationships and then inter-sample dependencies for the final diagnosis.}
\label{fig:fig3}
\end{figure*}

Graph Neural Networks (GNNs) \cite{RN179,RN180,RN194} offer a powerful framework for relational reasoning. The common GNN-based paradigm (conceptually illustrated in Fig. \ref{fig:fig2}) constructs a graph from pre-extracted features for fusion and analysis \cite{RN181,RN183}. While this approach has achieved notable success, we argue its efficacy for handling the unique complexities of multimodal medical data is curtailed by three critical limitations:

\textbf{1) Indiscriminate feature representation:} Existing methods often conflate diagnostically relevant features with redundant or noisy information, which can in turn impair the model's downstream reasoning process.
\textbf{2) Static graph topology:} They typically rely on a single, predefined graph structure, which is inherently insufficient to capture the diverse and patient-specific relationships within complex multimodal data.
\textbf{3) Localized message passing:} Most GNNs confine message passing to local neighborhoods, which restricts their ability to capture crucial long-range dependencies and form a holistic, global understanding of the data.

Thus, we propose the Multi-Activation Plane Interaction Graph Neural Network (MAPI-GNN), which shifts the paradigm from a single, static graph to learning a dynamic, multifaceted graph profile per patient, systematically responding to the three aforementioned limitations. Specifically: 
\textbf{(1)} To mitigate \textbf{indiscriminate feature representation}, a Multi-Dimensional Feature Discriminator (MDFD) adaptively identifies salient features via semantic importance. 
\textbf{(2)} To counter \textbf{static graph topology}, these salient features guide a Multi-Activation Graph Construction Strategy (MAGCS) to build multiple, semantically-aware topologies. 
\textbf{(3)} Finally, to move beyond \textbf{localized message passing}, a Hierarchical Feature Dynamic Association Network (HFDAN) fuses the resulting intra-sample graphs and models inter-sample dependencies in a global graph for robust classification.
Our main contributions are three-fold:
\begin{itemize}
\item A feature-driven paradigm for dynamic graph construction, where topology is learned from semantic salience rather than being predefined, enabling the model to capture more adaptive, patient-specific relationships.
\item A hierarchical architecture that distills intra-sample insights from a manifold of activation graphs and contextualizes them within a global graph, enabling comprehensive and high-fidelity patient-level analysis.
\item Comprehensive experimental validation on two diverse medical datasets (multi-parametric MRI, and CT with clinical data), demonstrating that our work achieves state-of-the-art performance and robust generalizability.
\end{itemize}

\section{Related Work}
\label{sec:related}
\subsection{CNN-based Multimodal Fusion}
Effective feature fusion is central to multimodal medical diagnosis \cite{RN195, RN221}. Many approaches are CNN-based (Fig. \ref{fig:fig2}(a)), traditionally employing early fusion \cite{RN185, RN186}, which is sensitive to noise and misalignment, or late fusion, which combines predictions from modality-specific models but consequently misses crucial low-level correlations \cite{RN188, RN189}. While more advanced hybrid strategies, sometimes employing attention mechanisms, exist, their reliance on fixed, grid-based operations fundamentally limits their ability to effectively discern salient information from heterogeneous sources \cite{RN191, RN197}. To address this limitation, our work introduces a feature discriminator to guide a more discerning and effective fusion process.

\subsection{GNN-based Multimodal Fusion}
GNN-based classification (Fig. \ref{fig:fig2}(b)) offers a more powerful paradigm for relational modeling \cite{RN190, RN198}. Current methods are typically applied at two scales. In the context of medical diagnostics, these scales correspond to distinct analytical granularities, from localized feature analysis to holistic patient-profile assessment. While \textbf{node-level} approaches excel at localized tasks \cite{RN213, RN214, RN182, RN212}, they struggle to form a holistic, patient-level diagnosis due to their inherently limited receptive fields. In contrast, \textbf{graph-level} methods can analyze a patient's entire profile \cite{RN215, RN211}, sometimes using multiplex structures \cite{RN193, RN184}. However, their core limitation is the reliance on predefined, static graph topologies, which implicitly assume that a single relational structure is optimal for all patients. This rigidity prevents the model from adapting to patient-specific pathological markers. To mitigate this bottleneck, our work proposes a dynamic graph construction strategy and a hierarchical fusion network to capture both adaptive intra-sample and global inter-sample relationships.

\begin{figure}[t]
\centering
\includegraphics[width=1.0\linewidth]{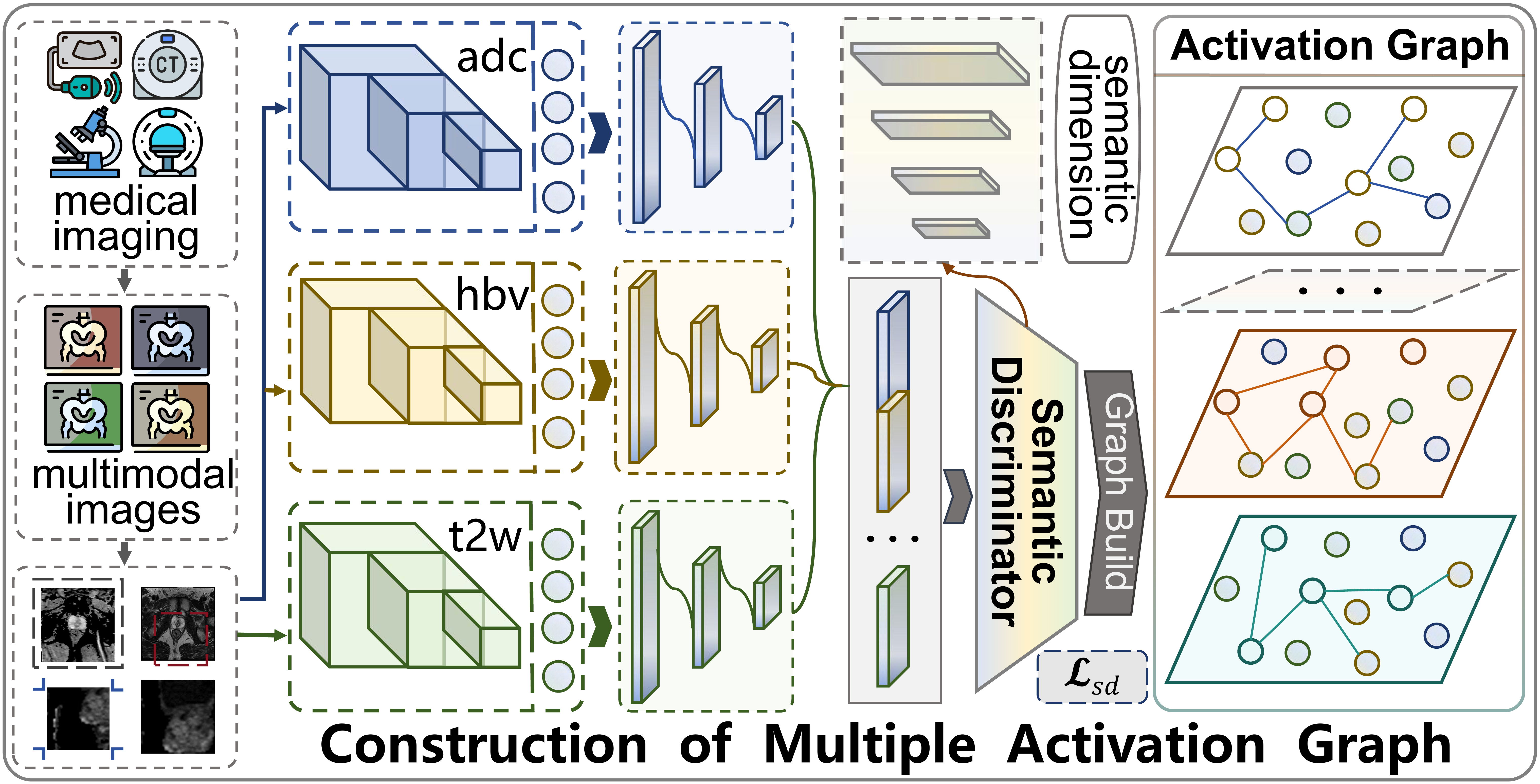}
\caption{Workflow of Stage I: Multi-Activation Graph Construction. From compressed modality features, a Multi-Dimensional Feature Discriminator identifies salient activated features for multiple semantic dimensions, each guiding the construction of a corresponding activation graph.}
\label{fig:fig4}
\end{figure}

\section{Method}
\label{sec:method}
\subsection{Overall Architecture}
Our proposed two-stage framework (Fig. \ref{fig:fig3}) counters the static single-graph paradigm by learning a patient-specific, multifaceted graph profile. First, a stack of semantically-aware activation graphs is dynamically constructed from disentangled feature subspaces. This profile is then hierarchically aggregated, modeling both intra- and inter-sample relationships to yield a robust final diagnosis.

\subsection{Stage I: Multi-Activation Graph Construction}
This stage (Fig. \ref{fig:fig4}) generates structured graphs from raw features by identifying discriminative feature subsets and using them to define graph topologies.

\subsubsection{Multi-Dimensional Feature Discriminator}
Our Multi-Dimensional Feature Discriminator assesses feature importance across multiple learned semantic dimensions. Initial modality features, extracted and compressed via autoencoders, are concatenated into a vector $\mathbf{x}\in \mathbb{R}^{C}$. This vector is projected by the Multi-Dimensional Feature Discriminator, $F_{sd}:\mathbb{R}^C\to\mathbb{R}^M$, into an $M$-dimensional semantic space. The influence of feature $i$ on semantic dimension $m$, $\mathbf{C}_m(i)$, is quantified via perturbation:
\begin{equation}C_m(i)=\left|[F_{sd}(\mathbf{x})]_m-[F_{sd}(\hat{\mathbf{x}}^{(i)})]_m\right|\end{equation}
where $\hat{\mathbf{x}}^{(i)}$ is $\mathbf{x}$ with $i$-th feature ablated. For each dimension $m$, features with the highest influence scores are designated as activated features.
To ensure the discriminator learns robust, disentangled semantic dimensions, training is guided by a composite loss, $\mathcal{L}_{sd}$, combining a primary reconstruction loss ($\mathcal{L}_{AE}$) with a regularization term $\mathcal{L}_{reg}(\Theta_{sd})$:
\begin{equation}\mathcal{L}_{sd}=\mathcal{L}_{AE}(\mathbf{x},\hat{\mathbf{x}})+\mathcal{L}_{reg}(\Theta_{sd})\end{equation}
where, $\mathcal{L}_{reg}$ enforces desirable properties on the discriminator's weights $\Theta_{sd}$ by combining L1 and L2 regularization, and an orthogonality constraint on its linear layers $\lambda_{orth}\|W_{sd}W_{sd}^T-I\|_F^2$ to promote semantic independence. Here, $|| \cdot ||_F$ denotes the Frobenius norm and $\lambda_{orth}$ is its corresponding weight coefficient.

\subsubsection{Multi-Activation Graph Construction Strategy}
The Multi-Activation Graph Construction Strategy builds a unique activation graph $\mathcal{G}_{m}$ for each semantic dimension $m$. All $M$ graphs share a common set of $C$ nodes $\mathcal{V}$, corresponding to the feature dimensions of $\mathbf{x}$. For each graph $\mathcal{G}_{m}$, the edge set $\mathcal{E}_m$ connects activated nodes to their $k$-nearest activated neighbors (via Euclidean distance in the initial feature space). We propose a edge weighting scheme where the weight $w_{ij}^{(m)}$ is the average influence of the connected nodes:
\begin{equation}
w_{ij}^{(m)}=\frac{1}{2}(C_m(i)+C_m(j))
\label{eq:edge_weight}
\end{equation}
This yields a set of $M$ graphs, $\{\mathcal{G}_1,\ldots,\mathcal{G}_M\}$, each offering a unique and complementary semantic view of the feature relationships for a given patient.

\begin{figure}[t]
\centering
\includegraphics[width=1.0\linewidth]{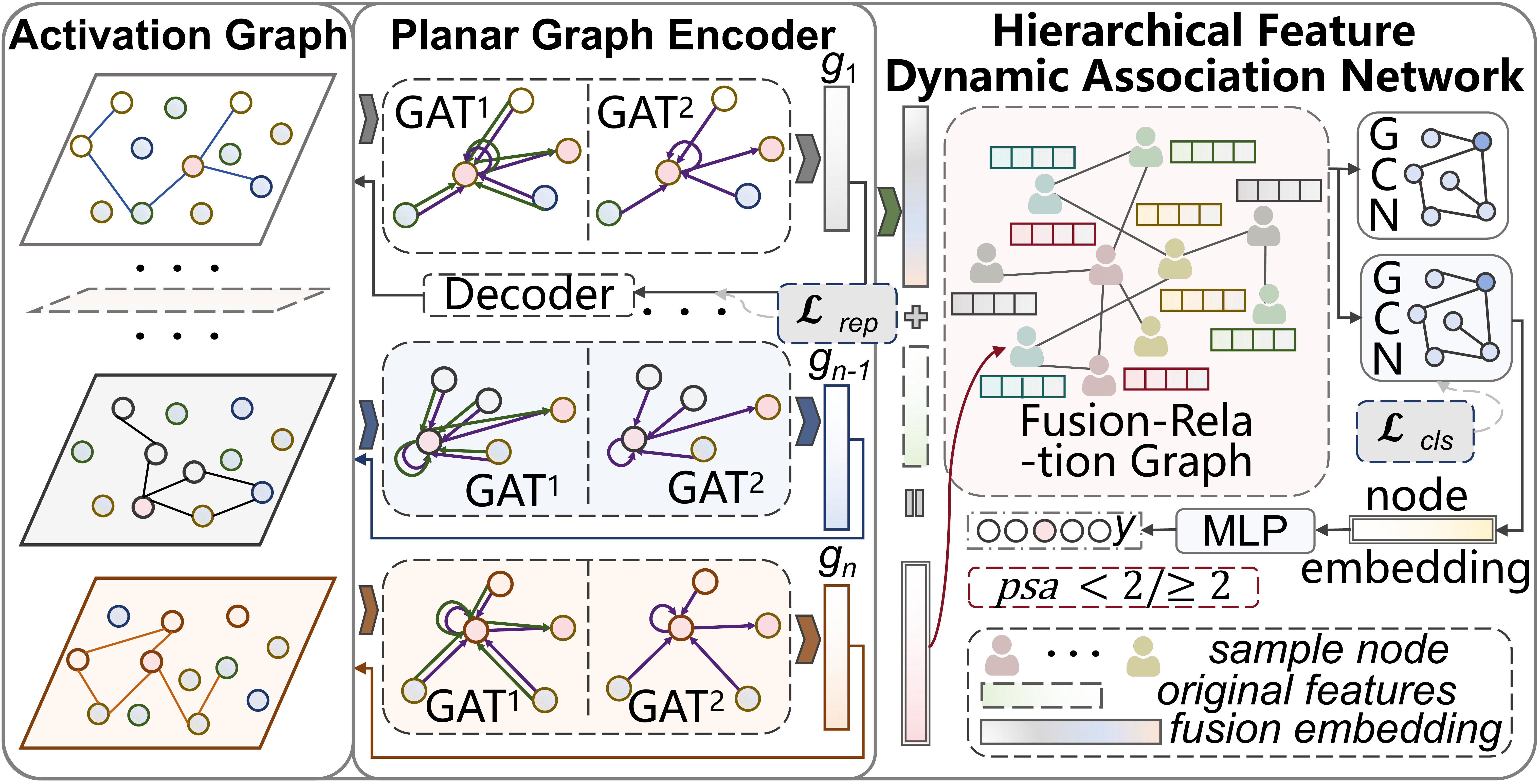}
\caption{Workflow of the Hierarchical Feature Dynamic Association Network (Stage II): 1) Intra-sample encoding of multiple activation graphs into representations ($\mathbf{g}_m$); and 2) Inter-sample classification on a global graph of fused patient vectors ($\mathbf{F}_p$) processed by a GCN.}
\label{fig:fig5}
\end{figure}

\begin{table*}[t]
\centering
\setlength{\tabcolsep}{16pt}
\begin{tabular}{l|c|c|c|c|c|c}
\hline
Method & ACC $\uparrow$ & AUC $\uparrow$ & PRE $\uparrow$ & REC $\uparrow$ & F1 $\uparrow$ & SPE \\
\hline
\multicolumn{7}{l}{\textit{CNN-based Methods}} \\
LFF & 0.8030 & 0.9605 & 0.9217 & 0.6364 & 0.7636 & \textbf{0.9697} \\
MSC & 0.8522 & 0.9380 & 0.8708 & 0.8181 & 0.8470 & 0.8863 \\
DenseNet & 0.7386 & 0.8078 & 0.7273 & 0.8333 & 0.7767 & 0.6250 \\
CNN & 0.7045 & 0.7867 & 0.7115 & 0.7708 & 0.7400 & 0.6250 \\
\hline
\multicolumn{7}{l}{\textit{GNN-based Methods}} \\
SAGE & 0.8068 & 0.8790 & 0.8650 & 0.7270 & 0.7900 & 0.8860 \\
GTAD & 0.8300 & 0.8200 & 0.7640 & \textbf{0.9550} & 0.8480 & 0.7850 \\
MGNN-CMSC & 0.8939 & 0.9592 & 0.8810 & 0.9487 & 0.9136 & 0.8148 \\
LG-GNN & 0.8977 & 0.9724 & 0.9149 & 0.8958 & 0.9053 & 0.9000 \\
HGM2R & 0.9242 & 0.9798 & 0.9246 & 0.9242 & 0.9242 & 0.9394 \\
\hline
\multicolumn{7}{l}{\textit{Transformer-based Methods}} \\
ViT & 0.9053 & 0.9728 & 0.8587 & 0.9491 & 0.9145 & 0.8069 \\
\hline
\multicolumn{7}{l}{\textit{Conventional Fusion Methods}} \\
Early-fusion & 0.7500 & 0.8255 & 0.7500 & 0.8125 & 0.7800 & 0.6750 \\
Late-fusion & 0.6591 & 0.6576 & 0.6667 & 0.7500 & 0.7059 & 0.5500 \\
\hline
\textbf{MAPI-GNN (Ours)} & \textbf{0.9432} & \textbf{0.9838} & \textbf{0.9361} & 0.9545 & \textbf{0.9438} & 0.9318 \\
\hline
\end{tabular}
\caption{Performance of different methods on the PI-CAI dataset. The best results are highlighted in \textbf{bold}.}
\label{tab:comparison}
\end{table*}

\subsection{Stage II: Hierarchical Feature Dynamic Association Network}
The Hierarchical Feature Dynamic Association Network (Fig. \ref{fig:fig5}) is tasked with processing the activation graphs from Stage I, hierarchically fusing their diverse representations to yield a final diagnosis.

\subsubsection{Intra-Sample Encoding and Fusion}
Each activation graph $\mathcal{G}_{m}$ is processed by a Planar Graph Encoder (implemented with GAT \cite{52}), yielding a 32-dimensional graph-level representation $\mathbf{g}_m$. Crucially, our pre-defined sparse topology $\mathcal{E}_m$ complements the GAT by constraining its attention mechanism to a small, semantically meaningful subset of feature relationships, guiding the model toward discriminative information. Furthermore, the learned attention coefficients are modulated by our pre-defined edge weights $w_{ij}^{(m)}$, ensuring that both learned patterns and a priori feature importance guide the final aggregation. The GAT layer aggregates neighbor information as:
\begin{equation}\mathbf{h}_i^{\prime}=\sigma\left(\sum_{j\in\mathcal{N}_i}\alpha_{ij}\mathbf{Wh}_j\right)\end{equation}
where $\alpha_{ij}$ is the learned attention coefficient and $\mathbf{h}_j$ is the node's representation. A readout function then produces the graph representation $\mathbf{g}_m=\frac{1}{C}\sum_{i=1}^C\mathbf{h}_i^{(L)}$, by averaging the final node representations from the $L$-th (final) layer.

These $M$ graph representations are concatenated with the initial feature vector $\mathbf{x}_p$ to form patient-level feature $\mathbf{F}_p$, preserving the complete feature profile from all $M$ semantic views and the original features in a lossless manner:
\begin{equation}
\mathbf{F}_p = \mathrm{Concat}(\mathbf{g}_1, \mathbf{g}_2, \dots, \mathbf{g}_M, \mathbf{x}_p)
\end{equation}

To ensure the learned embeddings are informative, the Planar Graph Encoders are regularized by a representation loss, $\mathcal{L}_{rep}$, which penalizes the reconstruction error when decoding initial node features from the final embeddings:
\begin{equation}\mathcal{L}_{rep}=\frac{1}{N\cdot M}\sum_{p=1}^N\sum_{m=1}^M\mathcal{L}_{MSE}\left(\mathbf{H}_{m,p}^{(0)},\mathrm{Decoder}(\mathbf{H}_{m,p}^{(L)})\right)\end{equation}
where $\mathbf{H}_{m,p}^{(0)}$ and $\mathbf{H}_{m,p}^{(L)}$ are the initial and final ($L$-th layer) node features for the $m$-th graph of the $p$-th patient.

\subsubsection{Inter-Sample Classification}
To model inter-sample relationships, we construct a global Fusion-Relation Graph, whose nodes are patients featured by $\mathbf{F}_p$. This global graph is processed by a Graph Convolutional Network (GCN \cite{51}), whose layer-wise propagation is:
\begin{equation}
\mathbf{H}^{(l+1)} = \sigma\left(\mathbf{\tilde{D}}^{-\frac{1}{2}}\mathbf{\tilde{A}}\mathbf{\tilde{D}}^{-\frac{1}{2}}\mathbf{H}^{(l)}\mathbf{W}^{(l)}\right)
\end{equation}
where $\mathbf{\tilde{A}}$ is the adjacency matrix with self-loops, $\mathbf{\tilde{D}}$ is diagonal degree matrix, and $\mathbf{H}^{(l)}$ is the $l$-th layer's representation.

The final representations are fed to an MLP for classification. This final stage is supervised by the primary task objective, the standard Cross-Entropy (CE) loss:
\begin{equation}
\mathcal{L}_{cls} = \mathcal{L}_{CE}(Y, \hat{Y})
\end{equation}
where $Y$ and $\hat{Y}$ are the ground-truth labels and the model predictions, respectively.

\subsection{Overall Training Objective}
The proposed architecture is jointly trained end-to-end by minimizing a final composite objective function. This objective is carefully formulated as a weighted sum of the aforementioned loss components to effectively balance their respective contributions:
\begin{equation}
\mathcal{L} = \lambda_{cls}\mathcal{L}_{cls} + \lambda_{rep}\mathcal{L}_{rep} + \lambda_{sd}\mathcal{L}_{sd}
\end{equation}
where $\lambda_{cls}$, $\lambda_{rep}$, and $\lambda_{sd}$ are hyperparameters to balance the losses. We set these to $\lambda_{cls}=1.0$, $\lambda_{rep}=0.3$, and $\lambda_{sd}=1.0$ to prioritize the main classification task and the semantic disentanglement, while applying moderate regularization to the representation encoding.

\section{Experiments}
\subsection{Datasets and Experimental Setup}
Our primary benchmark is \textbf{the public PI-CAI 2022 Challenge dataset}, used for clinically significant Prostate Cancer (csPCa) classification. The modalities for this task include T2w, ADC, and HBV MRI. We construct our study cohort from the 220 csPCa cases with trusted expert annotations, pairing them with 220 randomly under-sampled benign cases to form a balanced 440-case dataset. Preprocessing follows the prior study \cite{RN206}. To validate the robustness and generalizability of our proposed architectural components, we also utilize \textbf{a multi-modal Coronary Heart Disease (CHD) dataset} for diagnosis (974 cases, under IRB No. KY2025331). The modalities for this task are CCTA (Coronary CT Angiography) scans and structured clinical data. For a rigorous and reproducible evaluation, all models are assessed using a five-fold cross-validation protocol with a fixed random seed. We report the mean of key performance metrics, including Accuracy (ACC), Area Under the Curve (AUC), F1-Score (F1), Precision (PRE), Recall (REC), and Specificity (SPE).

\begin{figure}[t]
\centering
\includegraphics[width=1\columnwidth]{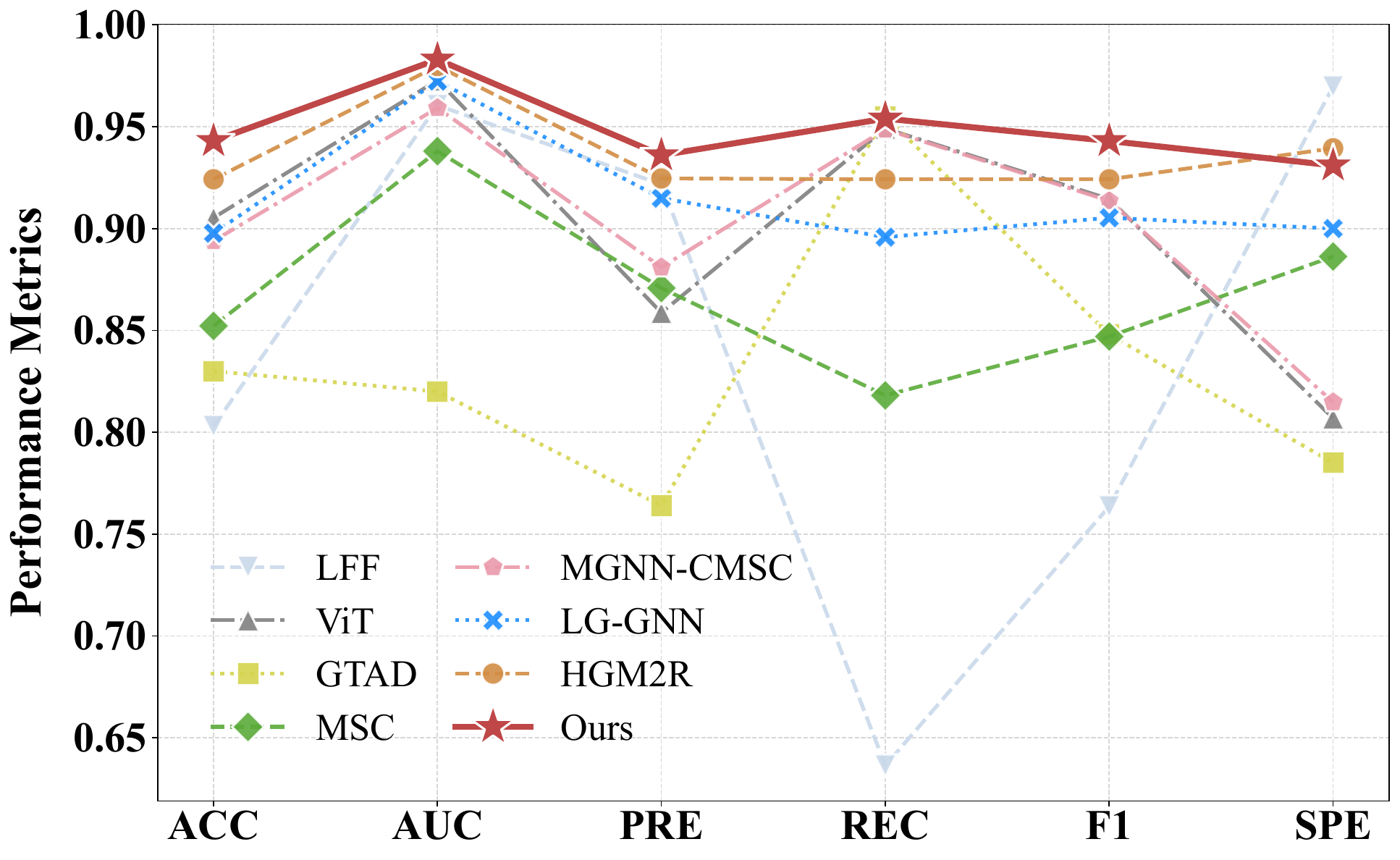}
\caption{Visual comparison of MAPI-GNN against key baselines on the PI-CAI dataset. This figure visualizes the primary metrics presented in Table~\ref{tab:comparison}, highlighting our model's competitive performance.}
\label{fig:fig6}
\end{figure}

\subsection{Comparison with State-of-the-art Methods}
\paragraph{Compared methods.}
To provide a comprehensive evaluation, we benchmark our proposed architecture against a diverse set of methods, spanning from foundational benchmarks to recent state-of-the-art (SOTA) approaches.
These are grouped into four categories: 1) \textbf{CNN-based methods}, representing advanced architectures for multimodal learning, such as LFF \cite{RN202} and MSC \cite{RN204}; 2) a wide range of \textbf{GNN-based methods} that constitute the current state-of-the-art in this domain, including GraphSAGE \cite{RN209}, GTAD \cite{RN217}, MGNN-CMSC \cite{RN218}, LG-GNN \cite{RN203}, and HGM2R \cite{RN216}; 3) \textbf{Transformer-based Methods}, such as a standard Vision Transformer (ViT); and 4) \textbf{conventional fusion strategies} (Early and Late fusion) which serve as fundamental benchmarks. To ensure a fair and reproducible comparison, all models were re-implemented or run using their official code under identical experimental settings.

\paragraph{Results Analysis.}
The quantitative results in Table \ref{tab:comparison} demonstrate our model's state-of-the-art performance. A key observation is the general superiority of GNN-based methods over their CNN-based counterparts. For instance, the strongest GNN baseline (HGM2R, 0.9242 ACC) outperforms the best-performing CNN (MSC, 0.8522 ACC) and ViT (0.9053 ACC) baselines, affirming the advantages of explicit relational modeling for this task. Building upon this paradigm, our framework further advances the SOTA by outperforming the highly competitive HGM2R with a notable margin of 1.9 percentage points in Accuracy (0.9432 vs. 0.9242). Crucially, our model's ability to achieve top scores across multiple key metrics simultaneously demonstrates a well-balanced and robust performance, rather than being narrowly optimized for a single metric. This dominant performance validates the synergistic effectiveness of our core design principles: dynamic, feature-driven graph construction and our hierarchical fusion architecture.

\begin{table}[t]
\centering
\setlength{\tabcolsep}{4pt}
\begin{tabular}{l|c|c|c|c|c}
\hline
Method & ACC & AUC & PRE & REC & F1 \\
\hline
w/o MDFD & 0.8500 & 0.9137 & 0.8515 & 0.8636 & 0.8533 \\
w/o MAGCS & 0.8205 & 0.9115 & 0.8024 & 0.8545 & 0.8266 \\
w/o HFDAN & 0.8364 & 0.9153 & 0.8234 & 0.8591 & 0.8402 \\
\textbf{MAPI-GNN} & \textbf{0.9432} & \textbf{0.9838} & \textbf{0.9361} & \textbf{0.9545} & \textbf{0.9438} \\
\hline
\end{tabular}
\caption{Ablation study on the PI-CAI dataset. Removing each of our three core components (MDFD, MAGCS, HFDAN) validates their respective contributions.}
\label{tab:ablation}
\end{table}

\begin{figure}[t]
\centering
\includegraphics[width=1\columnwidth]{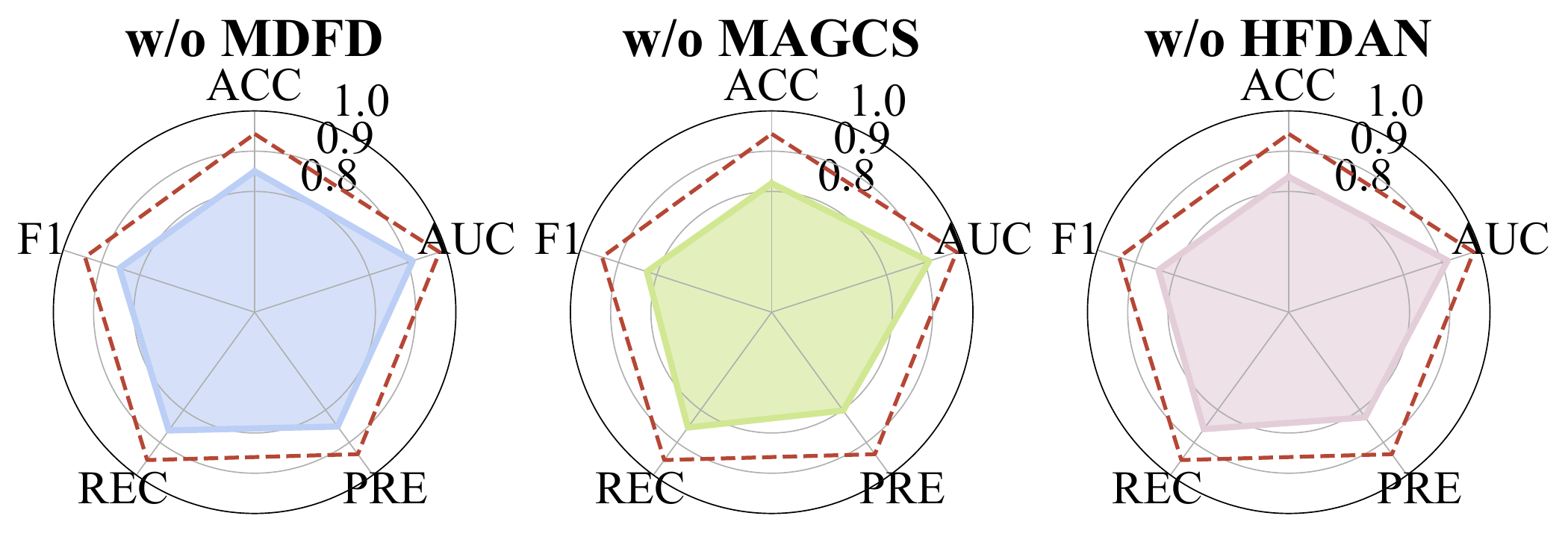}
\caption{Radar chart visualization of the PI-CAI ablation study. The reduced area upon removing any single component visually confirms that all modules are essential.}
\label{fig:fig7}
\end{figure}

\begin{table}[t]
\centering
\setlength{\tabcolsep}{4pt}
\begin{tabular}{l|c|c|c|c|c}
\hline
Method & ACC & AUC & PRE & REC & F1 \\
\hline
w/o MDFD & 0.8333 & 0.8672 & 0.7722 & 0.8839 & 0.8652 \\
w/o MAGCS & 0.8673 & \textbf{0.9538} & 0.8852 & 0.8710 & 0.8780 \\
w/o HFDAN & 0.8584 & 0.9306 & \textbf{0.9245} & 0.8033 & 0.8596 \\
\textbf{MAPI-GNN} & \textbf{0.9027} & 0.9206 & 0.8806 & \textbf{0.9516} & \textbf{0.9147} \\
\hline
\end{tabular}
\caption{Ablation study on the CHD dataset.}
\label{tab:ablation_chd}
\end{table}

\begin{figure}[t]
\centering
\includegraphics[width=1.0\columnwidth]{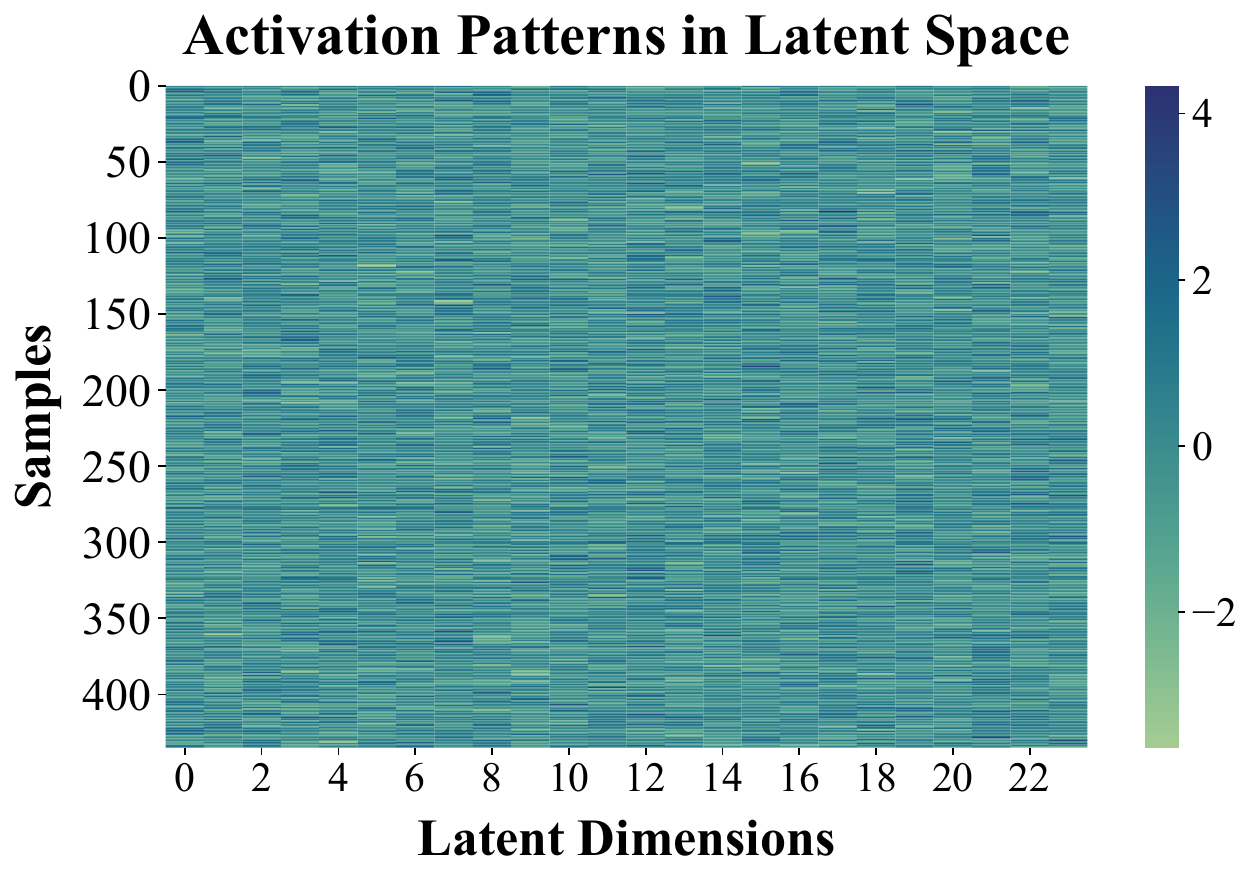} 
\caption{Heatmap of the MDFD latent space. It confirms the discriminator learns diverse, disentangled representations, as different semantic dimensions (columns) show distinct activation patterns across the patient samples (rows).}
\label{fig:heatmap}
\end{figure}

\begin{figure}[t]
\centering
\includegraphics[width=1\columnwidth]{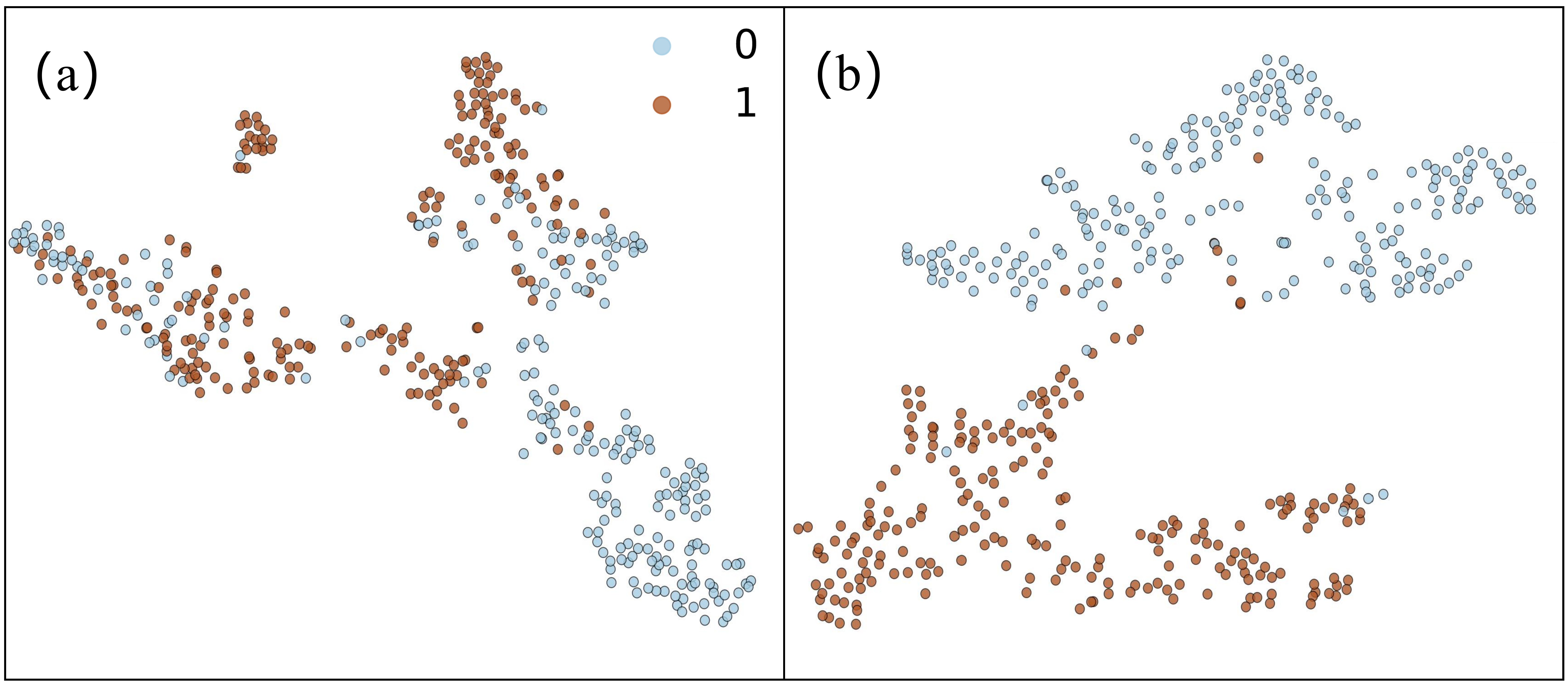} 
\caption{t-SNE visualization. (a) The original feature space, showing highly entangled classes. (b) After MDFD projection, the learned representations show clear separability, validating the creation of a discriminative embedding space.}
\label{fig:tsne01}
\end{figure}

\begin{figure}[t]
\centering
\includegraphics[width=1\columnwidth]{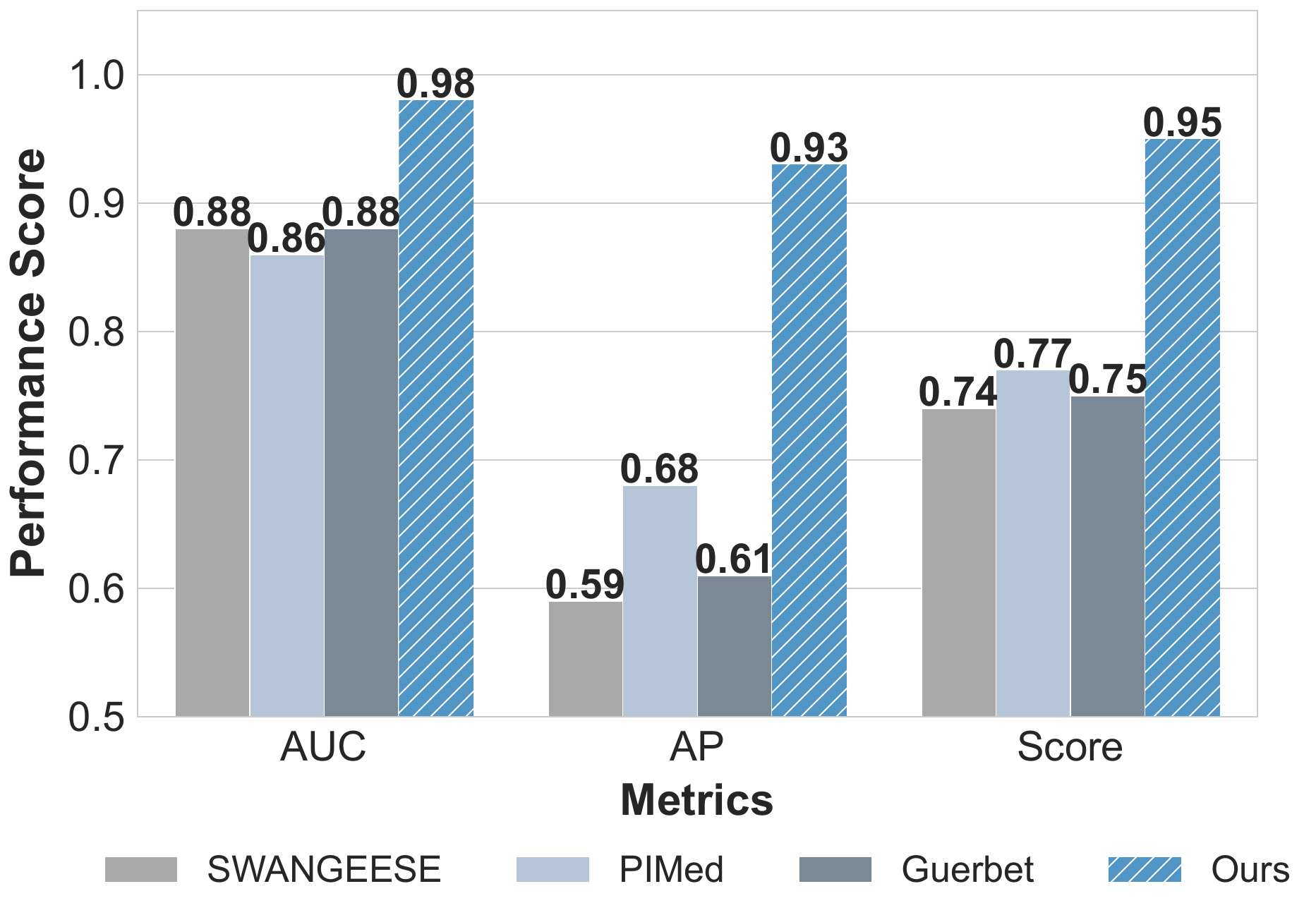}
\caption{Visual benchmark vs. top PI-CAI 2022 Challenge teams. Our model's scores, obtained on our cross-validation test fold, are compared against the leaderboard scores.}
\label{fig:fig8}
\end{figure}

\begin{table}[t]
\centering
\setlength{\tabcolsep}{7pt}
\begin{tabular}{l|c|c|c}
\hline
Team & AUC & AP & SCORE \\
\hline
SWANGEESE Team & 0.8860 & 0.5930 & 0.7400 \\
PIMed Team & 0.8650 & 0.6810 & 0.7730 \\
Guerbet Research & 0.8890 & 0.6150 & 0.7520 \\
\hline
\textbf{Ours} & \textbf{0.9838} & \textbf{0.9361} & \textbf{0.9599} \\
\hline
\end{tabular}
\caption{Benchmark against top teams from the PI-CAI 2022 Challenge leaderboard (SCORE = (AUC + AP) / 2).}
\label{tab:challenge}
\end{table}

\begin{figure}[t]
\centering
\includegraphics[width=1\columnwidth]{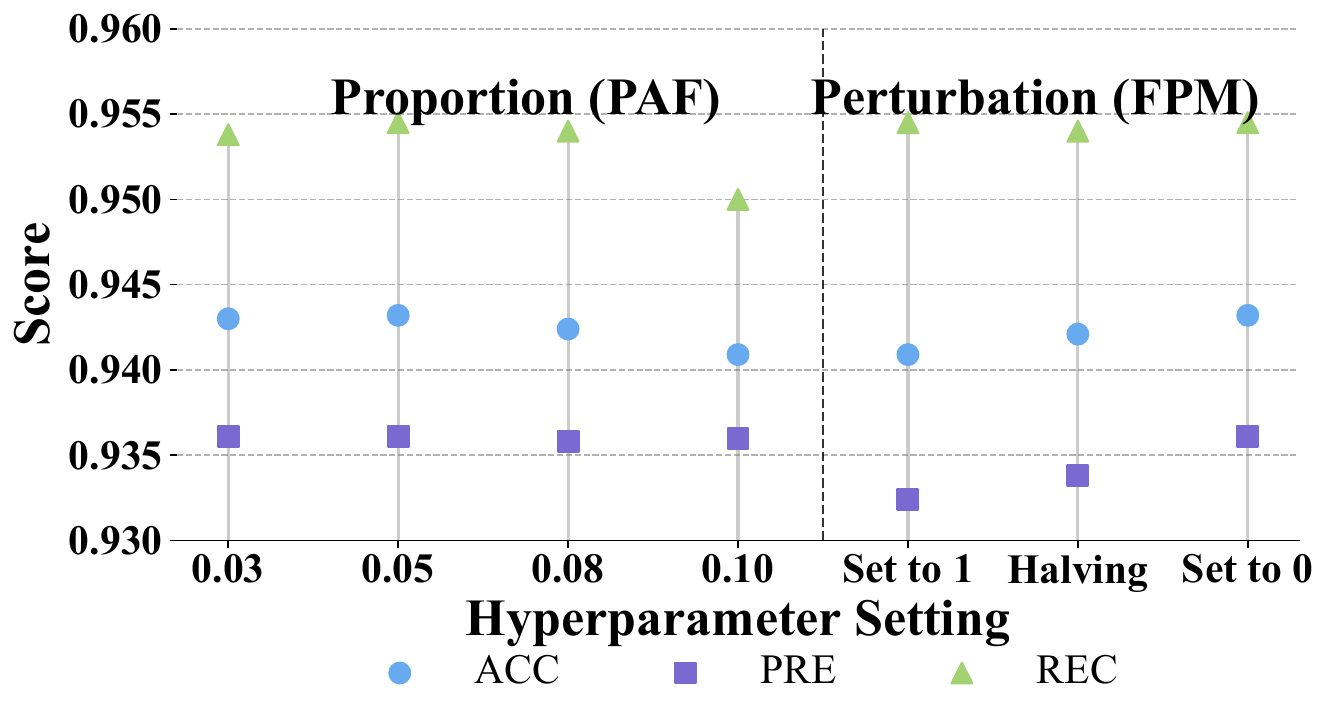}
\caption{Hyperparameter sensitivity analysis on the PI-CAI dataset. PAF: Proportion of Activated Features, FPM: Feature Perturbation Method.}
\label{fig:fig9}
\end{figure}

\begin{table}[t]
\centering
\setlength{\tabcolsep}{8pt}
\begin{tabular}{l|c|c|c}
\hline
Proportion (PAF) & ACC & PRE & REC \\
\hline
0.03 & 0.9430 & 0.9361 & 0.9538 \\
0.10 & 0.9409 & 0.9360 & 0.9500 \\
0.08 & 0.9424 & 0.9358 & 0.9540 \\
\textbf{0.05} & \textbf{0.9432} & \textbf{0.9361} & \textbf{0.9545} \\
\hline
\end{tabular}
\caption{Sensitivity to the Proportion of Activated Features (PAF). This value controls graph sparsity, and an optimal trade-off is achieved at 5\%.}
\label{tab:param_paf}
\end{table}

\begin{table}[t]
\centering
\setlength{\tabcolsep}{8pt}
\begin{tabular}{l|c|c|c}
\hline
Perturbation Method & ACC & PRE & REC \\
\hline
Set to 1 & 0.9409 & 0.9324 & 0.9545 \\
Halving & 0.9421 & 0.9338 & 0.9540 \\
\textbf{Zeroing-out} & \textbf{0.9432} & \textbf{0.9361} & \textbf{0.9545} \\
\hline
\end{tabular}
\caption{Comparison of different feature perturbation methods (FPM). The zeroing-out strategy used in our model yields the best performance.}
\label{tab:param_pert}
\end{table}

\subsection{Ablation Studies}
We conducted ablation studies on the PI-CAI and CHD datasets by systematically removing each core component: the Multi-Dimensional Feature Discriminator (MDFD), the Multi-Activation Graph Construction Strategy (MAGCS), and the Hierarchical Feature Dynamic Association Network (HFDAN). The results (Table~\ref{tab:ablation}, Table~\ref{tab:ablation_chd}, Fig.~\ref{fig:fig7}) confirm that all modules are integral, as removing any component causes a notable performance degradation across both datasets.
Interestingly, component importance varies by data modality. On the PI-CAI (mpMRI) dataset, removing MAGCS incurred the most significant accuracy drop (12.3\%), suggesting that capturing diverse inter-feature relationships is paramount. In contrast, on the CHD (heterogeneous CT and clinical) dataset, removing MDFD was most detrimental (6.9\% drop), highlighting the need for salient feature selection.
Furthermore, while some ablated models on the CHD dataset excel in specific metrics (e.g., AUC or PRE), our full model secures the best overall ACC and F1-Score (Table~\ref{tab:ablation_chd}). This demonstrates that the synergistic interplay of our components, rather than any single module, is key to achieving robust, adaptable performance.

\subsection{Qualitative Analysis}
To illustrate how our Multi-Dimensional Feature Discriminator (MDFD) transforms the feature space, we visualize its impact via t-SNE (Fig.~\ref{fig:tsne01}). The original features (Fig.~\ref{fig:tsne01}(a)) show highly overlapping csPCa and benign classes, indicating poor separability. In stark contrast, the representations processed by the MDFD (Fig.~\ref{fig:tsne01}(b)) exhibit clear class separation. This demonstrates that the MDFD successfully projects entangled raw features into a highly discriminative embedding space. Furthermore, the heatmap of this learned space (Fig.~\ref{fig:heatmap}) reveals diverse activation patterns across samples, suggesting the capture of rich, multifaceted abstract features. Together, these visualizations confirm the module's fundamental role in enhancing feature discriminability.

\subsection{Comparison with PI-CAI Challenge Leaders}
To situate our model's performance in a competitive context, we benchmark MAPI-GNN against the results of top-performing teams from the PI-CAI 2022 Challenge, including the SWANGEESE Team \cite{RN206}, PIMed Team \cite{RN205}, and Guerbet Research \cite{RN201}. As the official submission channel is closed and test labels are private, a direct leaderboard comparison is impossible. We strictly followed the official protocol to provide the most rigorous benchmark possible, training on the public data and reporting performance on our cross-validation test fold using the official metrics (AUC, AP, and SCORE = (AUC + AP) / 2). As shown in Table~\ref{tab:challenge} and Fig.~\ref{fig:fig8}, our architecture significantly surpasses these leading solutions, achieving a state-of-the-art SCORE of 0.9599 (AUC 0.9838, AP 0.9361). While this comparison is across different test sets, the substantial performance margin strongly suggests our method's architectural superiority and clinical potential.

\subsection{Parameter Analysis}
We analyze the sensitivity of our work to several key hyperparameters on the PI-CAI dataset to validate our final model configuration. Results are visualized in Fig.~\ref{fig:fig9}.

\paragraph{Number of Semantic Dimensions ($M$).}
This parameter defines the number of activation graphs. We analyzed its impact on the PI-CAI dataset: accuracy rose from 0.9136 ($M$=8) to 0.9302 ($M$=12) and plateaued around 0.9400 ($M$=16, 20), peaking at 0.9432 for $M$=24. We selected $M$=24 as the optimal trade-off between marginal performance gain and computational cost.

\paragraph{Graph Construction Parameters ($k$ and $w$).}
For the number of neighbors ($k$), testing $k \in \{3, 5, 10\}$ yielded accuracies of 0.9091, 0.9432, and 0.9205, respectively. We confirmed $k=5$ as the optimal trade-off between performance and cost. For edge weighting (Eq. \ref{eq:edge_weight}), we chose the average form. This links weights to discriminator scores ($C_m$) to guide GAT's attention without distorting individual scores.

\paragraph{Proportion of Activated Features (PAF).} This value controls the sparsity of the constructed graphs, creating a critical trade-off between retaining sufficient signal information (higher PAF) and filtering out potential noise (lower PAF). The results in Table~\ref{tab:param_paf} indicate that selecting 5\% (PAF=0.05) of features provides the optimal balance.

\paragraph{Feature Perturbation Method (FPM).} We also validate our choice of the zeroing-out perturbation strategy for calculating the crucial feature influence scores. As detailed in Table~\ref{tab:param_pert}, this method proves most effective for accurately identifying critical diagnostic features compared to other strategies (e.g., halving or setting to one).

\subsection{Computational Complexity}
To assess clinical feasibility, we analyzed MAPI-GNN's computational footprint on a single NVIDIA A30 GPU (165W TDP). Our model is lightweight, with 12.27M parameters and a total of 1.925 GFLOPs for the 440-case PI-CAI dataset, averaging 4.38 MFLOPs per case. With an average inference time of 45 ms per case, the model enables near real-time diagnostics. This high accuracy, fast inference, and MFLOPs-scale footprint confirm our framework as a practical and efficient solution for clinical deployment.

\section{Discussion} 
\paragraph{On the Synergy of CNNs and GNNs.} 
Our experiments affirm a key synergy. While GNNs alone outperform CNNs in relational modeling, our model demonstrates that using CNNs for potent feature extraction and GNNs to model their inter-dependencies yields superior performance to either architecture in isolation.

\paragraph{Clinical Applicability.} 
The architecture shows strong clinical relevance. Its high specificity (0.931) and recall (0.954) promise to reduce unnecessary biopsies while maintaining high detection rates. Furthermore, its efficient, end-to-end design enables streamlined inference, enhancing deployment feasibility over complex multi-stage methods.

\paragraph{Limitations and Future Work.} Despite its strong performance, we identify key avenues for future work. First, the current framework assumes complete modalities; enhancing its robustness to missing data is a critical next step for real-world utility. Second, while validated on two tasks, extending the framework to more diseases and data types (e.g., PET, histopathology, genomics) is needed to establish broader generalizability. While our analysis shows the learned semantic space is discriminative, developing methods, such as those linking to traditional radiomics, to map these abstract dimensions to concrete pathological concepts would be valuable towards greater clinical interpretability.

\section{Conclusion}
In this work, we introduce Multi-Activation Plane Interaction Graph Neural Network, a framework for multimodal medical diagnosis that, instead of relying on the prevailing static single-graph paradigm, learns patient-specific graph topologies directly from the data. At the core of our method is a two-stage process that first dynamically constructs a multifaceted graph profile for each patient and then performs a hierarchical fusion across these graphs to model both intra- and inter-sample relationships. Extensive experiments on diverse medical datasets demonstrate that our proposed framework achieves state-of-the-art performance. Ablation studies further confirm that its synergistic components are all integral to its success. By providing a more adaptive, powerful, and data-driven approach to multimodal fusion, our work represents a significant step toward more accurate, reliable, and ultimately more trustworthy computer-aided diagnosis in complex clinical scenarios.

\section{Acknowledgments}
This work was supported by the National Natural Science Foundation of China under Grant 62401481, Natural Science Foundation of Sichuan Province under Grant 2025ZNSFSC1450, Fundamental Research Funds for the Central Universities under Grant 2682024CX067, China Postdoctoral Science Foundation under Grant 2024M752683.

\bibliography{aaai2026}

\end{document}